%% file: main.tex
\documentclass[10pt,twocolumn,letterpaper]{article}

\usepackage[pagenumbers]{iccv} % To force page numbers, e.g. for an arXiv version

\input{preamble}

\usepackage{graphicx}
\usepackage{caption}
\usepackage{stfloats}
\usepackage{algorithm}
\usepackage{algpseudocode}
\usepackage{amsmath}
\usepackage[accsupp]{axessibility}

\definecolor{iccvblue}{rgb}{0.21,0.49,0.74}
\usepackage[pagebackref,breaklinks,colorlinks,allcolors=iccvblue]{hyperref}

\def\paperID{12181} % *** Enter the Paper ID here
\def\confName{ICCV}
\def\confYear{2025}

\title{PINO: Person-Interaction Noise Optimization for Long-Duration and Customizable Motion Generation of Arbitrary-Sized Groups}

\author{
  Sakuya Ota$^{1}$\footnotemark[1]\qquad
  Qing Yu$^{2}$\footnotemark[1]\ \footnotemark[2]\qquad
  Kent Fujiwara$^{2}$\qquad
  Satoshi Ikehata$^{3}$\qquad
  Ikuro Sato$^{1}$\\[0.35em]
  $^{1}$Institute of Science Tokyo\qquad
  $^{2}$LY Corporation\qquad
  $^{3}$National Institute of Informatics (NII)\\[0.3em]
  {\small\url{https://sinc865.github.io/pino/}}
}

\begin{document}

\twocolumn[{
    \maketitle
    
    \vspace{-1.5em}
    
    \centering
    
    \includegraphics[width=1\linewidth]{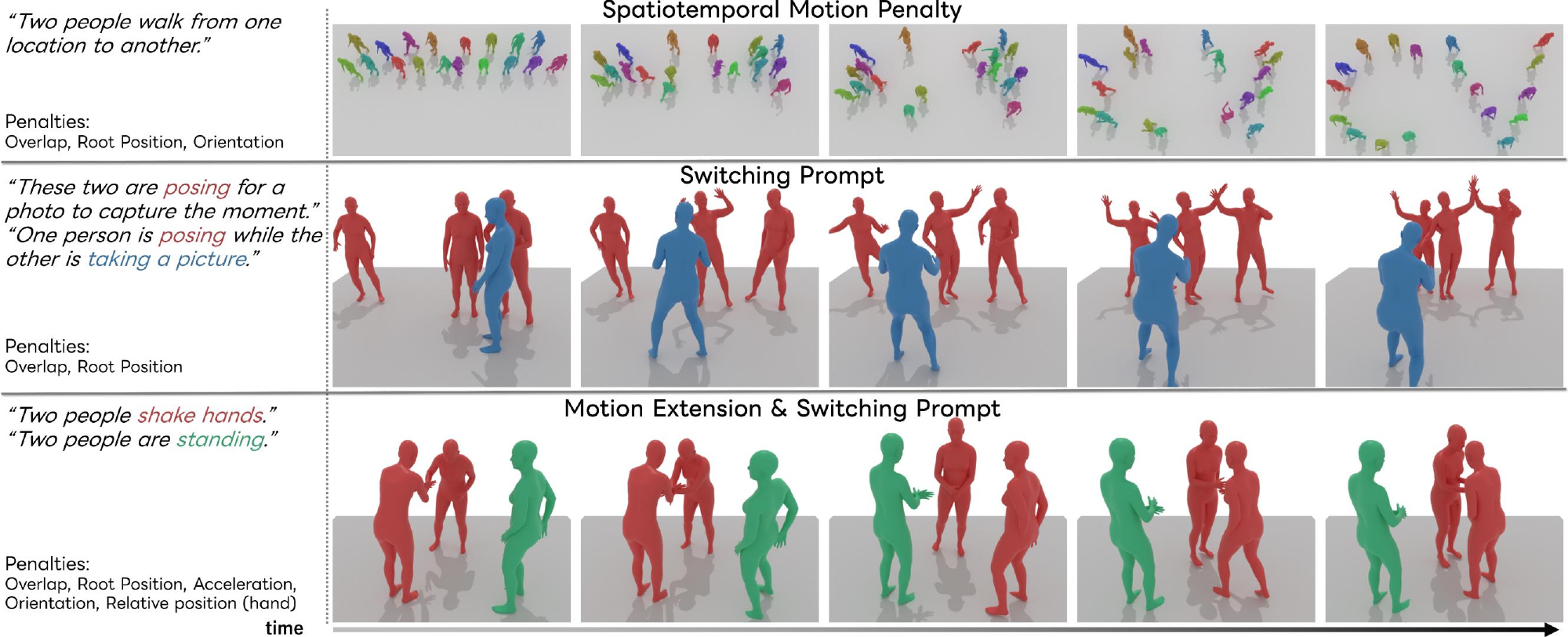}
    
    \captionsetup{type=figure, hypcap=true}
    \caption{\textbf{Person-Interaction Noise Optimization (PINO)} leverages pre-trained two-person interaction diffusion models as  
    motion priors to generate motions of arbitrary-sized groups 
    with the combination of spatiotemporal penalty, prompt switching between pairs, and motion extension. The first example employs penalties to guide humans into the form ``CV'', the second generates a pair posing for a photograph, then switches the prompt to generate the cameraman, and the third generates an alternating handshake sequence between three individuals by extending  motions while switching the prompt between the two pairs.} 

    \label{fig:method}

    \vspace{1em}
}]

\begingroup
  \renewcommand\thefootnote{\fnsymbol{footnote}}
  \footnotetext[1]{Equal contribution.}
  \footnotetext[2]{Corresponding author, email: \href{mailto:yu.qing@lycorp.co.jp}{yu.qing@lycorp.co.jp}}
\endgroup

\input{sec/0_abstract}    
\input{sec/1_intro}
\input{sec/2_relatedwork}
\input{sec/3_preliminaries}

\input{sec/4_method}

\input{sec/5_experiments}
\input{sec/6_conclusion}

%-------------- Acknowledgements --------------
\section*{Acknowledgements}
This study was carried out using the TSUBAME4.0 supercomputer at Institute of Science Tokyo.
%----------------------------------------------

{
    \small
    \bibliographystyle{ieeenat_fullname}
    \bibliography{main}
}

\input{sec/X_suppl}

\end{document}

%% file: preamble.tex
\newcommand{\Tref}[1]{Table~\ref{#1}}

\newcommand{\Fref}[1]{Fig.~\ref{#1}}

\newcommand{\Aref}[1]{Algorithm~\ref{#1}}

\newcommand{\x}{\mathbf{x}}
\newcommand{\X}{\mathbf{X}}

%% file: sec/0_abstract.tex
\begin{abstract}
Generating realistic group interactions involving multiple characters remains challenging due to increasing complexity as group size expands. While existing conditional diffusion models incrementally generate motions by conditioning on previously generated characters, they rely on single shared prompts, limiting nuanced control and leading to overly simplified interactions. In this paper, we introduce Person-Interaction Noise Optimization (PINO), a novel, training-free framework designed for generating realistic and customizable interactions among groups of arbitrary size. PINO decomposes complex group interactions into semantically relevant pairwise interactions, and leverages pretrained two-person interaction diffusion models to incrementally compose group interactions. To ensure physical plausibility and avoid common artifacts such as overlapping or penetration between characters, PINO employs physics-based penalties during noise optimization. This approach allows precise user control over character orientation, speed, and spatial relationships without additional training. Comprehensive evaluations demonstrate that PINO generates visually realistic, physically coherent, and adaptable multi-person interactions suitable for diverse animation, gaming, and robotics applications.

\end{abstract}

%% file: sec/1_intro.tex
\section{Introduction}
\label{sec:intro}
Generating realistic interactions among multiple characters is essential in animation, gaming, and robotics, where natural, coordinated behaviors enhance immersion and storytelling. While significant progress has been made in generating single-person motions~\cite{guo2022generating, tevet2023mdm, zhang2024motiondiffuse, chen2023executing} and two-person interactions~\cite{liang2024intergen, tanaka2023interaction, shafir2024priormdm}, extending these approaches to interactions involving three or more people poses challenges.

A straightforward approach trains models on annotated motions from larger groups~\cite{shan2024multiperson}. However, as the number of characters increases, the interactions become exponentially more complex, rendering accurate textual annotation and simultaneous motion generation intractable.

To generate arbitrary-sized group interactions, InterControl~\cite{wang2023intercontrol} and FreeMotion~\cite{fan2024freemotion} have introduced conditional motion diffusion models based on ControlNet~\cite{zhang2023adding}. These models generate new motion for each character incrementally by conditioning it on the motions of previously generated characters. This enables the generation of one character at a time and requires training only on one- or two-person datasets, thereby mitigating scalability issues.

While these models generate each motion individually, group-level interactions are generated using a single shared prompt, despite the inherent difficulty of describing interactions among many characters compared to those for one or two individuals. Moreover, conditioning the generation of new characters on the entire history of existing ones prevents a clear mapping between each character and its specific role in the prompt. Consequently, the demonstrated group interactions tend to be uniform and overly simplified (\eg, ``three people walking in a circle''~\cite{fan2024freemotion}, ``Three people are holding hands together''~\cite{wang2023intercontrol}). While these methods provide basic guidance, such as collision avoidance~\cite{wang2023intercontrol} or approximate positioning~\cite{fan2024freemotion}, they lack fine-grained motion customization. Achieving precise control would require re-training ControlNet for each control signal.

In this paper, we introduce a training‐free framework for generating group interactions of arbitrary size with customizable user controls. Our key insight is that group interactions consist of smaller interconnected interactions, where characters interact directly and serve as pivots linking multiple interactions. For example, a group photo is not a single monolithic interaction, but a composite of interactions between the photographer and each poser, as well as between the posers themselves. Even when the overall group action appears uniform (\eg, ``three people walking in a circle''), it can be broken down into smaller interactions (\eg, between the first and second individuals, and between the second and third), underscoring the role of shared characters in driving group dynamics.

Building on this insight, we propose Person-Interaction Noise Optimization (PINO). Unlike previous methods that generate a group interaction from a shared prompt~\cite{shan2024multiperson, wang2023intercontrol, fan2024freemotion}, PINO decomposes the task into sequential pairwise interaction generations, enabling each interaction to be guided by a novel, semantically relevant text prompt. This modular approach leverages pretrained diffusion models for two-person interactions~\cite{liang2024intergen} and incrementally expands them into a cohesive group dynamic. For instance, in the group photo scenario shown in~\cref{fig:method}, our method first generates the interaction between the photographer and one subject using a prompt (\eg, ``one person is posing while the other is taking a picture''). As additional characters join, each new interaction is paired with an existing character (either the photographer or a subject) using prompts such as ``these two are posing for a photo to capture the moment,'' thereby anchoring the overall group interaction.

A common challenge when naïvely composing group interactions from pairwise interactions is the emergence of unrealistic dynamics, such as overlapping poses or body penetration among characters. To address this, PINO enforces physical plausibility by optimizing the initial noise in the diffusion model with physics-based penalties inspired by recent diffusion noise optimization techniques~\cite{karunratanakul2023dno, tevet2023mdm, chefer2023attend, guo2024initno}. This optimization evaluates the states of all individuals and imposes penalties based on root position, movement region, orientation, and relative positioning, significantly reducing artifacts and ensuring that the generated motions are both physically plausible and contextually coherent.

Our post-hoc noise optimization provides extensive user control over character direction, speed, and spatial relationships, enabling precise customization for diverse applications without additional training (unlike~\cite{wang2023intercontrol, fan2024freemotion}). Furthermore, by incorporating motion inpainting techniques, PINO can generate long-duration group interactions with timing-specific text prompts (see~\cref{fig:method}-bottom). 

Quantitative and qualitative evaluations both demonstrate that PINO achieves visual realism, physical plausibility, and adaptability, by generating natural, coherent group interaction sequences without requiring dedicated multi-person training data.

%% file: sec/2_relatedwork.tex
% \begin{figure*}
%   \centering
%   \includegraphics[width=0.90\linewidth]{figs/fig2v4.pdf}
%   \caption{Overview of the PINO framework. PINO takes the motion of the existing individual as the reference motion and generates the new individual through sequential motion composition. Then, PINO optimizes the initial latent noise of the new added motion by applying various penalties in order to achieve coherence and avoid inconsistencies with the existing motions.}
%   \label{fig:short}
% \end{figure*}

\section{Related Work}
\label{sec:relatedwork}

\subsection{Single-Person Motion Generation}
 % With the development of motion capture methods and equipments, analysis of skeletal human motion has garnered a wide range of attention.
 Advances in motion-capture systems have spurred broad interest in skeletal motion analysis. One central topic is human motion generation for single-person scenarios~\cite{yan2019convolutional,zhao2020bayesian,holden2020learned}. Early works tackled label-conditioned motion generation through the use of variational autoencoders (VAE)~\cite{petrovich2021actor}, generative adversarial networks (GAN)~\cite{xu2023actformer}, and diffusion models~\cite{ho2020denoising}. Recently, with the rise of language models, text-conditioned motion generation~\cite{athanasiou2022teach, chen2023executing, kalakonda2022action, petrovich2022temos, tevet2023mdm, zhang2023t2m, zhang2023remodiffuse, dai2024motionlcm, guo2024momask, barquero2024seamless, pinyoanuntapong2024mmm, ghosh2021synthesis} has emerged as a prominent field of research. Diffusion models including MDM~\cite{tevet2023mdm}, MotionDiffuse~\cite{zhang2024motiondiffuse}, and MLD~\cite{chen2023executing} excel in generating high-quality, text-conditioned motions by using embeddings from pre-trained text encoders such as CLIP~\cite{clip}.
 
 In addition, some works of research 
 focus on task-specific motion generation, including control over specific joint positions~\cite{shafir2024priormdm, karunratanakul2023gmd, xie2024omnicontrol} and creation of long-duration motion sequences~\cite{tseng2023edge, zhang2023diffcollage, shafir2024priormdm, lee2024t2lm, barquero2024seamless}. 
 However, these %works 
 are primarily limited to single-person motion, with minimal focus on multi-person 
 interactions.

\subsection{Multi-Person Interaction Generation}
%Generating motions for multiple individuals is more complex than single-person motion due to dynamic interactions among individuals.
Generating motions for multiple individuals is more complex than for a single person due to dynamic interactions among them.
Most existing works focus on two-person interactions. ComMDM~\cite{shafir2024priormdm} was an early attempt at two-person generation by fine-tuning a single-person model, though performance was limited by data scarcity. %RIG~\cite{tanaka2023interaction} used predefined roles and text descriptions, while InterGen~\cite{liang2024intergen} introduced a large-scale dataset of two-people interactions named InterHuman and a shared-weight diffusion framework with cross-attention. 
RIG~\cite{tanaka2023interaction} incorporated role-specific text prompts, while InterGen~\cite{liang2024intergen} introduced InterHuman, a large-scale two-person dataset, along with a shared-weight diffusion model using cross-attention.
InterMask~\cite{javed2024intermask} applied VQ-VAE and transformers to model detailed 3D interactions. %As data is key to improving quality, there are also efforts to create larger
% dataset~\cite{xu2024inter}.
Larger datasets are being developed to enhance motion quality~\cite{xu2024inter}.

Recent studies have begun exploring generation of multi-person interactions beyond pairs. %InterControl~\cite{wang2023intercontrol} adapted a single-person motion model to multi-person scenarios by guiding joint positions for specific interactions, but its lack of training on multi-person data makes it difficult to achieve interactive elements such as synchronicity and timings.
%interaction clarity.
InterControl~\cite{wang2023intercontrol} guided joint positions using a single-person model, but since it was not trained on multi-person data,  it struggles to capture interactive features such as synchronicity.
FreeMotion~\cite{fan2024freemotion}, inspired by ControlNet~\cite{zhang2023adding}, treated individual character motions as control signals, allowing sequential generation of text-conditioned multi-person interactions. However, because the method does not apply any constraint on the generated motions, artifacts such as body penetration emerge as character count increases. %However, a lack of diverse multi-person data led to artifacts such as body penetration. %as character count increased. 

Shan et al.~\cite{shan2024multiperson} proposed a Transformer-based diffusion model trained on a new multi-person dataset, which first generates multi-person poses from text and then produces corresponding motion sequences. This approach requires a dedicated multi-person dataset, which is difficult to gather for varying group sizes. Additionally, generation is restricted to predefined group sizes, with a single text prompt to describe the group action, causing lack of flexibility and control over the motion of each individual. %Our approach avoids the issues of these methods by utilizing a two-person network and applying various penalties to generate realistic interactions between multiple individuals. 

%However, this approach lacked scalability, restricting the generation to predefined group sizes.

\subsection{Noise Optimization for Generation}
% In the context of image generation with diffusion models, there have been attempts to improve the alignment between the input text prompt and the output image by finding a better latent noise.

For diffusion-based image generation, several works search for optimal latent noise to better align text prompts and outputs. Chefer~\etal~\cite{chefer2023attend}, attempted to achieve this goal by slightly editing the latent noise at each timestep. Guo~\etal~\cite{guo2024initno} improved upon this by solving for the optimal noise based on scores that evaluate attention maps.

Recent studies in human motion generation have also explored noise optimization during inference. Approaches like ProgMoGen~\cite{liu2024programmable} and DNO~\cite{karunratanakul2023dno} optimize the initial noise of MDM~\cite{tevet2023mdm} based on differentiable objectives, guiding the diffusion process to generate motions aligned with specific goals without retraining the model.

Our work introduces Person-Interaction Noise Optimization (PINO) for multi-person interaction generation, which optimizes the initial noise of a pre-trained  two-person diffusion model %to tackle scalability challenges and 
to address limitations such as flexibility in terms of group size, motion length and additional control, all without requiring extra training nor dedicated multi-person datasets. 
We summarize the differences between the proposed method and related multi-person models in~\Tref{tab:related}.

% \begin{table}[t]
% \begin{center}
% \tabcolsep = 1mm
% \resizebox{0.99\linewidth}{!}{
% \begin{tabular}{c|c|c|c|c}
%     \toprule
%      Method & \begin{tabular}{c} \#Person in \\ Network \\ Design \end{tabular} & \begin{tabular}{c} \#Person in \\ Training\\ Data \end{tabular}  & \begin{tabular}{c} Order of \\ Multi-Person  \\Generation \end{tabular} & \begin{tabular}{c} Method for \\ Interaction \end{tabular} \\   \midrule
%      InterControl~\cite{wang2023intercontrol} & One & One  &  Sequential & ControlNet\\
%      FreeMotion~\cite{fan2024freemotion} & One & Two  &  Sequential & ControlNet\\
%      Shan et al.~\cite{shan2024multiperson} & Multiple & Multiple  &  Concurrent & \begin{tabular}{c} Supervised \\ Training \end{tabular} \\
%      Ours & \begin{tabular}{c} Two \\ (or multiple) \end{tabular}& Training-free  &  Sequential & \begin{tabular}{c} Noise \\ Optimization  \end{tabular} \\
%     \bottomrule
% \end{tabular}
% }
% \end{center}
% \vspace{-10pt}
% \caption{Summary of recent related methods for multi-person interaction generation.}
% \vspace{-5pt}
% \label{tab:related}
% \end{table}

\begin{table}[t]
\begin{center}
\tabcolsep = 1mm
\resizebox{0.99\linewidth}{!}{
\begin{tabular}{c|c|c|c|c}
    \toprule
     Method & \begin{tabular}{c} \#Person in \\ Network \\ Design \end{tabular} & \begin{tabular}{c} \#Person in \\ Training\\ Data \end{tabular}  & \begin{tabular}{c} Order of \\ Multi-Person  \\ Generation \end{tabular} & \begin{tabular}{c} Method \end{tabular} \\   \midrule
     InterControl~\cite{wang2023intercontrol} & One & One  &  Sequential & ControlNet\\
     FreeMotion~\cite{fan2024freemotion} & One & Two  &  Sequential & ControlNet\\
     Shan et al.~\cite{shan2024multiperson} & Multiple & Multiple  &  Concurrent & \begin{tabular}{c} Supervised \\ Training \end{tabular} \\
      ProgMoGen~\cite{liu2024programmable} & One & Training-free & \begin{tabular}{c}N/A  \\ (Single-Person)\end{tabular}& \begin{tabular}{c} Noise \\ Optimization \end{tabular} \\
     Ours & \begin{tabular}{c} Two \\ (or multiple) \end{tabular}& Training-free  &  Sequential & \begin{tabular}{c} Noise \\ Optimization  \end{tabular} \\

    \bottomrule
\end{tabular}
}
\end{center}
\vspace{-10pt}
\caption{Comparison of recent multi-person interaction generation methods, including single-person method ProgMoGen~\cite{liu2024programmable}.}
\vspace{-5pt}
\label{tab:related}
\end{table}

%% file: sec/3_preliminaries.tex
\section{Preliminaries}

\subsection{Multi-Person Motion Representation}

We denote the motion sequences of $M$ individuals as $\X = \{\x^1, \x^2, \dots, \x^M\}$, with $\x^p \in \mathbb{R}^{N \times D}$ containing $N$-frame, $D$-dimensional motion representations of the $p$-th individual.
For single-person motion generation, the representation of HumanML3D~\cite{guo2022generating} is commonly used. However, its normalization process removes relative positional information, which is essential for spatial interactions. %in multi-person scenarios. 
To address this limitation, we follow the non-canonical representation in InterGen~\cite{liang2024intergen} that retains global positional information, enabling better modeling of interactions among individuals.

The pose of individual $p$ at frame $n$ is defined as:
\begin{equation}
\x^p(n) = \left[ \mathbf{j}_\text{glo}^p,\ \mathbf{j}_\text{vel}^p,\ \mathbf{j}_\text{rot}^p,\ \mathbf{g}_\text{foot}^p \right],
\end{equation}
where $\mathbf{j}_\text{glo}^p \in \mathbb{R}^{3J}$ represents the global joint positions, $\mathbf{j}_\text{vel}^p \in \mathbb{R}^{3J}$ denotes the joint velocities, $\mathbf{j}_\text{rot}^p \in \mathbb{R}^{6J}$ is the local joint rotations, and $\mathbf{g}_\text{foot}^p \in \mathbb{R}^{4}$ represents binary foot-ground contact features. Here, $J$ is the number of joints. This formulation retains global coordinates, which are crucial for capturing the spatial relationships and interaction dynamics between individuals in multi-person motion.

\subsection{Diffusion-based Pairwise Motion Generation}
\label{sec:diff}
Our framework builds upon pretrained text-driven two-person motion generation diffusion models~\cite{liang2024intergen,tanaka2023interaction,Ruiz-Ponce_2024_CVPR}. While our approach is model-agnostic, we assume that the model can generate two-person motion from corresponding noisy motion vectors, $\x_T^1$ and $\x_T^2$, with a text prompt $c$, as:  

\begin{equation}  
\label{eq:inter}  
[\x_0^1,\, \x_0^2] = G_\theta(\x_T^1, \x_T^2, c).  
\end{equation}  
Here, $\x_0^1, \x_0^2 \in \mathbb{R}^{N \times D}$ represent the denoised motion sequences of the two individuals and $G_\theta$ is the base diffusion model, which predicts the noise at each time step and denoises the noisy vectors to recover the original motion sequence. 
Throughout this paper, we assume $G_\theta$ to be InterGen~\cite{liang2024intergen} with a DDIM noise sampler~\cite{song2021denoising}. InterGen, trained on the large-scale, text-annotated InterHuman dataset, generates realistic two-person motion by leveraging an interaction diffusion framework with shared-weight cooperative networks and mutual attention. For details on InterGen, we refer readers to the original paper~\cite{liang2024intergen}.

%% file: sec/4_method.tex
\section{Method} 
\begin{figure*}
  \centering
  \includegraphics[width=0.95\linewidth]{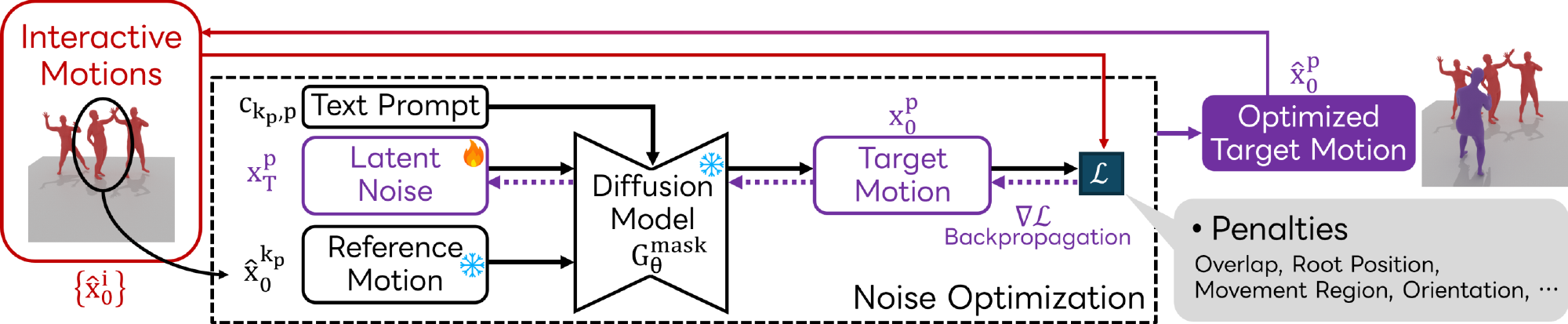}
  \caption{\textbf{Person-Interaction Noise Optimization (PINO).}
           \textcolor[HTML]{C00000}{\textbf{Red}} marks the already-generated reference motions, whereas 
           \textcolor[HTML]{7030A0}{\textbf{purple}} highlights the elements for the newly added person.
           A masked diffusion model transforms the latent noise of a new individual, which is guided by a text prompt and reference motion, into an initial motion that is subsequently refined through noise optimization. This process is then repeated for groups of arbitrary size.
           }
  \label{fig:short}
\end{figure*}

As depicted in~\Fref{fig:short}, PINO incrementally introduces new individuals through sequential motion composition. Each additional character is generated by conditioning on the motion of an existing individual as a pivot and subsequently refined using a series of penalty-based latent noise optimizations, ensuring high-quality interactions within the entire group. This approach is particularly advantageous when creating scenes with varied roles—for example, a dancing group contrasted with a set of onlookers.
\subsection{Input Specification for PINO}\label{sec:input}
The input to PINO consists of two main components:
\begin{enumerate}
\item \textbf{Participants}: An ordered list specifying the individuals involved in each pairwise interaction (target/reference).
\item \textbf{Interaction Descriptions}: Text prompts describing the interaction between each specified pair of participants.
\end{enumerate}
For example, in a scene involving a photographer and three subjects, the user could define the following pairings:
\begin{itemize}
\item Photographer (reference) $\leftrightarrow$ Subject A (target) \\
``One person is posing while the other is taking a picture.''
\item Subject A (reference) $\leftrightarrow$ Subject B (target) \\
``These two are posing for a photo to capture the moment.''
\item Subject B (reference) $\leftrightarrow$ Subject C (target) \\
``These two are posing for a photo to capture the moment.''
\end{itemize}
Each prompt is crafted based on the specific relationship between the conditioned character and the new character. This structured input guides the generation of coherent group interactions, offering a more detailed and controlled alternative to a single, unified prompt for the entire group.

\subsection{PINO: Person-Interaction Noise Optimization} 
\Aref{alg:multi_person_generation} summarizes our PINO approach, wherein each new individual is seamlessly integrated into the preceding interactions while preserving the physical and contextual coherence of the entire sequence.
\\\noindent
\textbf{First Interaction Generation}:\\
Given a list of participants and interaction descriptions as specified in~\cref{sec:input}, we initialize the first interaction between two participants as \(\x_0^1, \x_0^2\) with prompt \(c_{1,2}\) using the pre-trained two-person diffusion model \(G_\theta\):
\[
\x_0^1,\, \x_0^2 \leftarrow G_\theta(\x_T^1,\, \x_T^2, c_{1,2}).
\]
Since pre-trained two-person motion generation models often produce motions with artifacts such as overlapping poses or body penetration, we propose a joint pose refinement step. In this step, we optimize the initial noise \(\x_T^p\) of the diffusion process by minimizing a differentiable loss function that encodes the desired interaction criteria, and then regenerate the interaction:
\begin{eqnarray}
\hat{\x}_T^1 &\leftarrow& \arg\min_{\x_T^1} \mathcal{L}(G^{\rm mask}_\theta(\x_T^1, \x_0^2),\x_0^2,c_{1,2}),\label{eq:noise_opt1}\\
\hat{\x}_T^2 &\leftarrow& \arg\min_{\x_T^2} \mathcal{L}(G^{\rm mask}_\theta(\x_T^2, \x_0^1),\x_0^1,c_{1,2}),\label{eq:noise_opt2}\\
\hat{\x}_0^1, \hat{\x}_0^2 &\leftarrow& G_\theta(\hat{\x}_T^1,\, \hat{\x}_T^2, c_{1,2}).
\end{eqnarray}
Here, \(G_{\theta}^{\rm mask} (\x^{\rm tgt}, \x^{\rm cond}, c)\) is a variant of \(G_{\theta}\) modified to accept a pivotal condition in two-person generation. During the denoising process, the condition sequence \(\x^{\rm cond}_{t}\) for \(t \in \{0,\dots,T\}\) is replaced by a noisy version of the previously generated reference character motion \(\hat{\x}_0^{\rm cond}\), and only \(\x_T^{\rm tgt}\) is denoised. The loss function \(\mathcal{L}\) comprises cost terms that penalize physics-based artifacts, such as overlapping poses and body penetration. The details of the design of these cost functions will be discussed later. By back-propagating through the diffusion process, this optimization ensures that the generated motions \(\hat{\x}_0^1\) and \(\hat{\x}_0^2\) not only align with the text prompt but also maintain physical plausibility.
\\\noindent
\textbf{Group Interaction Generation}:\\
Given the first interaction $\{\hat{\x}_0^1,\, \hat{\x}_0^2\}$, we sequentially introduce additional characters \(\x_3, \dots, \x_N\) with corresponding prompts \(c_{k_3,3}, \dots, c_{k_p,p}, \dots, c_{k_N,N}\), where \(k_p\) is the reference character index as specified in the predefined interaction list. As illustrated in~\cref{fig:short}, for the \(p\)-th target character conditioned on its reference character with optimized noise \(\hat{\x}_0^{k_p}\) and prompt \(c_{k_p,p}\), we generate its motion as:
\begin{eqnarray}
\hat{\x}_T^p \leftarrow \arg\min_{\x_T^p} \mathcal{L}\Bigl(G^{\rm mask}_\theta(\x_T^p,\,\hat{\x}_0^{k_p},\, c_{k_p,p}),\, \{\x_0^{i \in \mathcal{I}}\}\Bigr),\label{eq:noise_opt}\\
\hat{\x}_0^p \leftarrow G^{\rm mask}_\theta(\hat{\x}_T^p,\, \hat{\x}_0^{k_p},\, c_{k_p,p}),
\end{eqnarray}
where \(\mathcal{I}\) denotes the predefined subset of previously generated individuals involved in the optimization. 
By fixing the motion of the reference character and conditioning the generation of the new character on it, we guarantee that the newly synthesized motion aligns with the intended interaction context. This optimization ensures that the generated motion \(\hat{\x}_0^p\) not only conforms to the reference but also harmonizes with the overall group dynamics.
This composition process can be applied recursively to incorporate additional characters—even across groups with distinct interaction contexts—by initializing new pairs when necessary.

\begin{algorithm}[t]
\caption{Pseudocode of PINO}
\label{alg:multi_person_generation}
\begin{algorithmic}[1]
\Require $M \geq 2$: Number of individuals
\Ensure $\mathbf{X}$: Multi-person interaction sequences
\State Initialize $\mathbf{X} \leftarrow \emptyset$, $\x_T^1, \x_T^2 \sim \mathcal{N}(\mathbf{0}, \mathbf{I})$
\State Generate motions $\x_0^1,\, \x_0^2 \leftarrow G_\theta(\x_T^1,\, \x_T^2, c_{1,2})$
\State Optimize $\hat{\x}_T^1 \leftarrow \arg\min_{\x_T^1} \mathcal{L}$ (Eq. \ref{eq:noise_opt1})
\State Optimize $\hat{\x}_T^2 \leftarrow \arg\min_{\x_T^2} \mathcal{L}$ (Eq. \ref{eq:noise_opt2})
\State Generate motions $\hat{\x}_0^1,\, \hat{\x}_0^2 \leftarrow G_\theta(\hat{\x}_T^1,\, \hat{\x}_T^2, c_{1,2})$
\State $\mathbf{X} \leftarrow \{\hat{\x}_0^1,\, \hat{\x}_0^2\}$

\For{$p = 3$ to $M$}
    \State Select individual with same context: \par$k \in \{1,\, 2,\, \dots,\, p-1\}$ 
    \State Initialize $\x_T^p \sim \mathcal{N}(\mathbf{0}, \mathbf{I})$
    % \State Generate motion $\x_0^p \leftarrow G_\theta(\hat{\x}_0^k,\, \x_T^p, c)$
    \State Select targeted individuals for interaction as \par$\mathcal{I} \subset \{1,\, 2,\, \dots,\, p-1\}$ 
    \State Optimize $\hat{\x}_T^p \leftarrow \arg\min_{\x_T^p} \mathcal{L}$ (Eq. \ref{eq:noise_opt})
    \State Generate motion $\hat{\x}_0^p \leftarrow G_\theta(\hat{\x}_0^k,\, \hat{\x}_T^p, c_{k_p,p})$
    \State $\mathbf{X} \leftarrow \mathbf{X} \cup \{\hat{\x}_0^p\}$
\EndFor
\State \textbf{return} $\mathbf{X}$
\end{algorithmic}
\end{algorithm}
\subsection{Optimization Penalties}
\textbf{Overlap Avoidance:}
During interaction generation, artifacts such as overlap between individuals can occur, leading to unrealistic results. This becomes prominent with more individuals, as the sequential motion composition strategy, by itself, generates 
 additional motions only by considering the context with the pivot individual. 
  To mitigate the issue, we introduce  
 $\mathcal{L}_{\text{overlap}}$ 
 that penalizes the root positions of individuals if they are closer than a predefined threshold $\delta$:
\begin{align}
\begin{split}
&\mathcal{L}_{\text{overlap}}(G^{\rm mask}_\theta(\x_T^p,\;\hat{\x}_0^{k_p},\;c_{k_p,p}), \{\hat{\x}_0^{i}\})  \\
&= \sum_{i}\sum_{n}  \max\left(0,\, \delta - \left\| \mathbf{p}^p_\text{root}(n) - \hat{\mathbf{p}}^i_\text{root}(n) \right\|_2 \right),
\end{split}
\end{align}
where $\mathbf{p}^p_\text{root}(n)$ and $\hat{\mathbf{p}}^i_\text{root}(n)$ are the root positions of the two individuals at frame $n$. 

\textbf{Spatio-Temporal Motion Control:}
To achieve controllable and customizable motion synthesis, 
we introduce spatio-temporal penalties 
$\mathcal{L}_{\text{control}}$, allowing users to 
guide motion generation. 
Controllable aspects include:

\begin{itemize}
    \item \textbf{Root Position Penalty}: Enforcing that an individual reaches a specific location at a particular time.
    \item \textbf{Movement Region Penalty}: Limiting individual movement within defined areas.
    \item \textbf{Orientation Penalty}: Controlling facing directions at specific frames.
    \item \textbf{Relative Position Penalty}: Maintaining desired distances or orientations between individuals.
\end{itemize}

These penalties are incorporated as differentiable loss functions, enabling optimization of the initial noise latent. For example, a root position constraint can be represented by the squared distance between the actual and target positions at a specific frame. We refer the readers to the supplementary material for the detailed definition of each penalty.  

By integrating these components, the overall optimization loss $\mathcal{L}$ for initial noise optimization is denoted as:
\begin{equation}
\mathcal{L} = \mathcal{L}_{\text{overlap}} + \mathcal{L}_{\text{control}}.
\end{equation}
This framework allows us to generate realistic, artifact-free motions that adhere to user-defined spatio-temporal requirements, providing a flexible and precise approach to motion customization of multi-person interactions.

\subsection{Long-Duration Interaction Generation} PINO is also designed to extend interactions over long durations, producing sequences that remain coherent and physically plausible. Starting with existing multi-person motion sequences ${\hat{\x}^i}$, we extract the final $n$ frames from each sequence to serve as a contextual reference. Then, through motion inpainting, subsequent frames are generated by integrating this reference at every denoising step: \begin{equation} \mathbf{x}_t^i \leftarrow \mathbf{m} \odot \hat{\x}^i + (1 - \mathbf{m}) \odot \mathbf{x}_t^i, \end{equation} where $t$ denotes the denoising timestep and $\mathbf{m}$ is a binary mask indicating the initial $n$ frames. To achieve smooth transition between reference and generated content, a boundary penalty is added to $\mathcal{L}$, which minimizes joint acceleration, thereby ensuring natural motion dynamics.

By uniting sequential motion composition, rigorous noise optimization, and advanced inpainting techniques, PINO offers a robust, training-free solution for generating natural, diverse, and controllable multi-person interactions. This framework not only overcomes the limitations of prior approaches but also provides extensive user control over the dynamics of group interactions.

%% file: sec/5_experiments.tex
\section{Experiments}

\begin{table}[t]
    \centering
    \begin{subtable}{1\linewidth}
        \centering
        \setlength{\tabcolsep}{1pt}
        \resizebox{\linewidth}{!}{
        \begin{tabular}{l|cccc}
            \toprule
            Method & Overlap $\downarrow$ & PenVol. (cm$^3$) $\downarrow$ & Foot Skate $\downarrow$ & Max Acc. $\downarrow$ \\
            \midrule
            GT & 0.029 & 471.75 & 0.099 & 0.074 \\
            \midrule
            InterGen & 0.119 & 3112.72 & \textbf{0.124} & \textbf{0.034} \\
            PINO-InterGen & \textbf{0.000} & \textbf{275.65} & 0.125 & \textbf{0.034} \\
            \bottomrule
        \end{tabular}
        }
        \caption{Non-semantic metrics.}
    \end{subtable}%
    \hfill
    \begin{subtable}{1\linewidth}
        \centering
        \setlength{\tabcolsep}{1pt}
        \resizebox{0.7\linewidth}{!}{
        \begin{tabular}{l|ccc}
            \toprule
            Method & FID $\downarrow$ & Div. $\rightarrow$ & R-Prec. $\uparrow$ (Top3) \\
            \midrule
            GT & 0.983 & 7.921 & 0.715 \\
            \midrule
            InterGen & 13.278 & 7.793 & 0.674 \\
            PINO-InterGen & \textbf{13.163} & \textbf{7.904} & \textbf{0.675} \\
            \bottomrule
        \end{tabular}
        }
        \caption{Semantic metrics.}
    \vspace{-10pt}
    \end{subtable}
    \caption{Evaluation of two-person interaction generation while avoiding overlap.}
    \vspace{-5pt}
    \label{tab:avoid_overlap}
\end{table}

\begin{table}[t]
    \centering
    \resizebox{\columnwidth}{!}{
        \begin{tabular}{lccccccc}
            \toprule
            Method &  Overlap↓ & PenVol.↓ & Foot Skate↓ & Max Acc.↓  \\
            \midrule
            InterGen (inpainting) &  0.766 & 25215.78 & 0.099 & \textbf{0.030}  \\
            PINO-InterGen &  \textbf{0.000} & \textbf{785.01} & \textbf{0.091} & 0.031 \\
            \bottomrule
        \end{tabular}
    }
    \vspace{-8pt}
    \caption{Evaluation of multi-person interaction generation}
    \vspace{-8pt}
    \label{tab:multi_person_comparison}
\end{table}

\begin{table}[t]
    \centering
    \resizebox{\columnwidth}{!}{
        \begin{tabular}{lccccc}
            \toprule
            Method & Pair & FID↓ & Overlap↓ & Foot Skate↓ & Max Acc.↓  \\
            \midrule
            FreeMotion & (1,2) & 19.383 & 0.255 & \textbf{0.112} & \textbf{0.034}\\
                     & (1,3) & 25.184 & 0.806 & \textbf{0.122} & 0.038\\
                     & (1,4) & 25.812 & 0.964 & 0.127 & 0.040\\
                     & (1,5) & 25.671 & 0.991 & 0.130 & 0.042\\
            \midrule
            InterGen & (1,2) & 13.100 & 0.119 & 0.126 & \textbf{0.034}\\
                     & (1,3) & 19.414 & 0.799 & 0.125 & \textbf{0.036}\\
                     & (1,4) & 19.386 & 0.937 & \textbf{0.124} & \textbf{0.038}\\
                     & (1,5) & 19.501 & 0.977 & \textbf{0.124} & \textbf{0.039}\\
            \midrule
            PINO-InterGen & (1,2) & \textbf{12.920} & \textbf{0.000} & 0.127 & \textbf{0.034} \\
                         & (1,3) & \textbf{17.170} & \textbf{0.002} & 0.135 &0.040 \\
                         & (1,4) & \textbf{16.890} & \textbf{0.019} & 0.138 &0.043 \\
                         & (1,5) & \textbf{16.911} & \textbf{0.069} & 0.142 & 0.045 \\
            \bottomrule
        \end{tabular}
    }
    \vspace{-7pt}
    \caption{Evaluation of multi-person interaction generation. From pair (1,2), interactions are generated incrementally using 1 as a pivot. Overlap is among all individuals generated up to that step.}
    \vspace{-7pt}
    \label{tab:pairwise_multi_person}
\end{table}

\subsection{Implementation Details}
In our experiments, we employ the official pre-trained weights of InterGen \cite{liang2024intergen}, which has been trained on the InterHuman dataset for two-person interactions. We keep these weights fixed throughout our process. For efficient latent noise optimization, we adopt a 50-step DDIM sampler. 
For consistency and fair comparison, all baseline methods are implemented using the same DDIM settings.

During the optimization phase, we set the learning rate to 0.003 and conduct 100 optimization steps to yield optimal results for the tasks. 
This setup effectively balances computational efficiency with the quality of the generated motions. Further details on the implementation, including specific hyperparameter choices and optimization strategies, are provided in the supplementary material.

\subsection{Multi-Person Interaction Generation}

\begin{figure*}[h]
    \centering
     \includegraphics[width=1\linewidth]{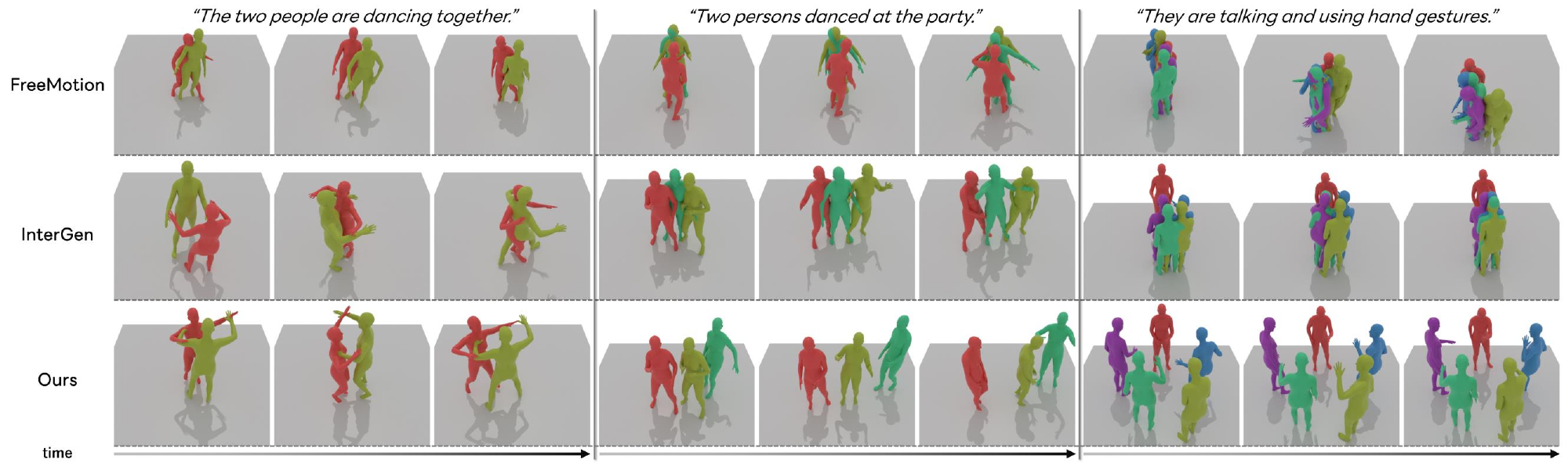}
    \vspace{-20pt}
    \caption{Comparison of the generated multi-person interaction.} 
     \vspace{-10pt}
    \label{fig:multi_person_comparison}
\end{figure*}

\begin{figure}[t]
    \centering
    \includegraphics[width=\linewidth]{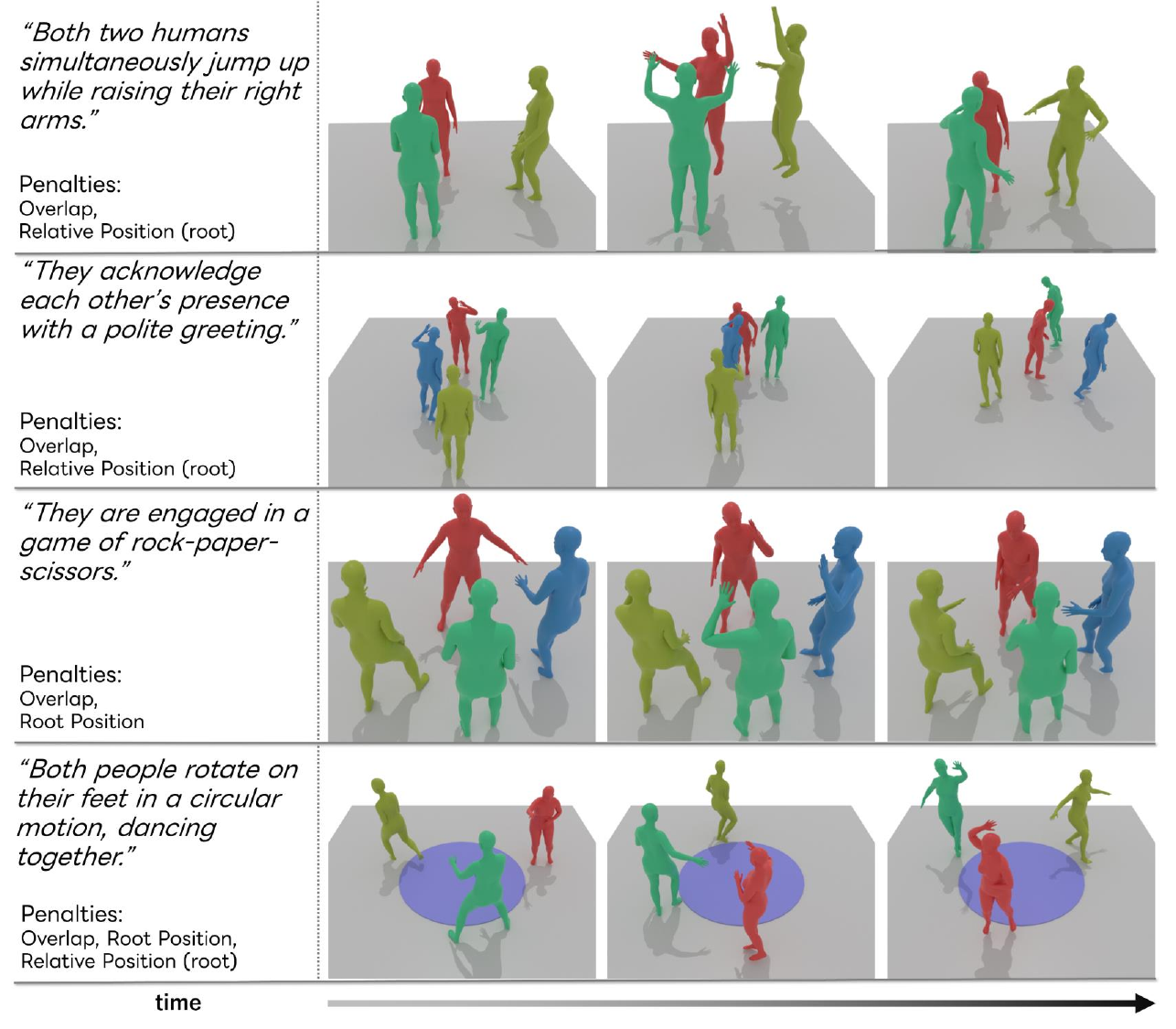}
    \vspace{-20pt}
    \caption{Qualitative results of multi-person interaction generation. Representative frames are shown for visualization.}
    \vspace{-10pt}
    \label{fig:example}
\end{figure}

\textbf{Evaluation Metrics:}
To quantitatively evaluate the quality of generated motions, we employ both non-semantic and semantic metrics. Non-semantic metrics include Overlap, Penetration Volume, Foot Skating, and Maximum Acceleration (Max Acc). Overlap is the probability that at least one pair of individuals 
comes closer than a predefined distance threshold at their root positions, with lower values reflecting more natural spacing, and thus, more realistic motion. 
Penetration Volume~\cite{lee2024interhandgen} evaluates physical plausibility by computing the maximum interpenetration volume across sequences using SMPL-fitted meshes~\cite{loper2015smpl}, instead of time-averaged values, to better capture sparse-contact interactions.
Foot Skating quantifies unnatural foot sliding to assess realism, while Max Acc. captures excessive accelerations that deviate from natural dynamics.

Semantic metrics include Fréchet Inception Distance (FID), Diversity, and R-Precision.
FID measures the distributional gap between generated and ground truth motions, where lower values denote a closer alignment with real data. Diversity captures variation across generated samples. R-Precision evaluates how well the generated motions match the semantics of real ones, with higher values indicating stronger alignment between text and motion.

\textbf{Two-Person Interaction:}
Due to scarcity of datasets that involve more than 3 individuals, we first evaluate the performance of PINO on two-person interaction generation task. We randomly select 300 samples from the InterHuman~\cite{liang2024intergen} test set to generate motions, and compare these quantitatively and qualitatively with the baseline, InterGen.

As can be seen from the results in Table~\ref{tab:avoid_overlap}, the base model InterGen contains artifacts, despite being trained on interaction data. 
Our method significantly improves Overlap and Penetration Volume, indicating minimal proximity issues and strong physical plausibility.
Semantic metrics confirm alignment with ground truth motion data, evidenced by low FID values. Diversity and R-precision scores further demonstrate that PINO is able to maintain the semantic accuracy of the base model, while avoiding artifacts such as collisions, also evident  in~\cref{fig:multi_person_comparison} (left).

\textbf{Multi-Person Interaction:}
We assess the plausibility of multi-person interactions by evaluating their physical and spatial consistency using non-semantic metrics. Given a simple text prompt, we first generate interactions between two individuals and then extend them to a third person. Specifically, we generate eight distinct two-person interactions from the same prompt, and then produce 12 variations of the motion of the third individual for each interaction. As shown in Table~\ref{tab:multi_person_comparison}, 
our PINO framework preserves the motion quality of the base model InterGen by providing similar values for Foot Skate and Max Acc., and at the same time mitigate undesirable physical artifacts, as shown in the metrics of Overlap and Penetration Volume. 

Additionally, we evaluate semantic correctness by decomposing each multi-person sequence into two-person pairs and computing FID, overlap, foot skating, and Max Acc. for every pair. In this evaluation, the baseline InterGen and the multi-person generator FreeMotion \cite{fan2024freemotion} serve as comparisons. For FreeMotion, previously generated characters are supplied as conditional inputs,  and new individuals are appended sequentially. To ensure fairness, we generate five-person motions with each method and evaluate only the four two-person pairs formed by the first character and each subsequently generated character.
Both InterGen and FreeMotion exhibit pronounced character overlap as the number of participants increases. Our PINO framework inherently sidesteps this issue, producing more coherent and realistic group interactions. While FreeMotion conditions each new character on all existing ones, it lacks post-generation relational constraints, limiting interaction fidelity. In contrast, PINO preserves motion quality while markedly reducing overlap and enhancing interaction realism.
Fig.~\ref{fig:multi_person_comparison} visualizes these qualitative gains, and Fig.~\ref{fig:example} shows additional examples of realistic multi-person interactions generated by our approach. Details of these experiments are provided in the supplementary material.

\subsection{Ablation Study on Motion Penalties}

\begin{figure}[t]
    \centering
    \includegraphics[width=\linewidth]{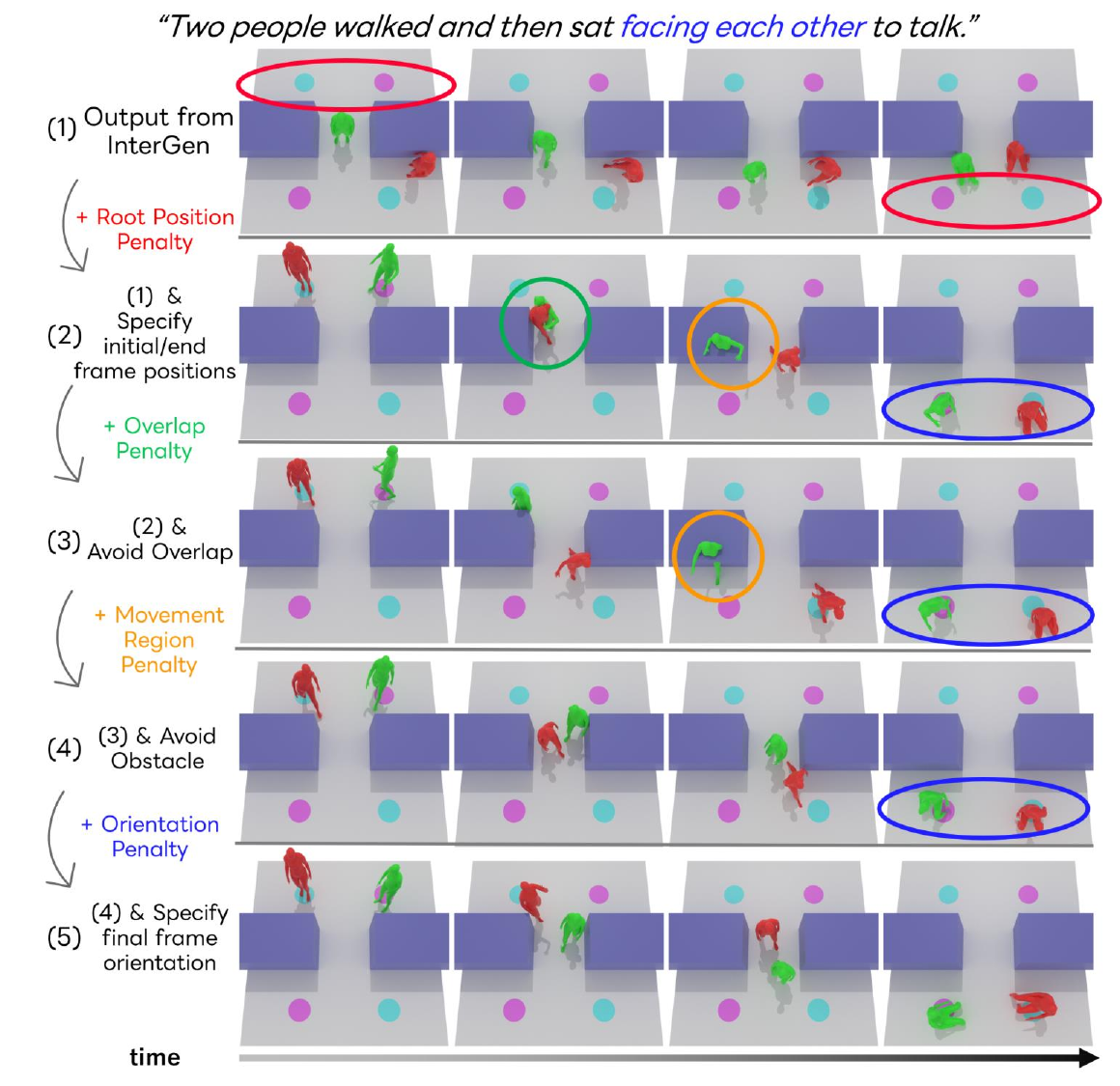}
    \vspace{-20pt}
    \caption{Visualization of the ablation study. We incrementally introduce the proposed penalties to InterGen (1). We highlight the unfulfilled demands: \textcolor{red}{Red circle}:  Individuals not at intended initial/end positions. \textcolor[HTML]{00B050}{Green circle}: Collision with other individual. \textcolor[HTML]{FE9802}{Orange circle}: Entering restricted regions. \textcolor{blue}{Blue circle}: Unintended orientation. By introducing additional penalties, our method is able to generate motions that fulfill the requests from the users and naturally fit in the provided context or environment.}
    \vspace{-5pt}
    \label{fig:ablation}
\end{figure}

We conduct an ablation study demonstrating how the penalty terms steer InterGen to satisfy external user demands. In the evaluation scene illustrated in Fig.~\ref{fig:ablation}, two agents follow the prompt, “Two people walked and then sat facing each other to talk,” moving from an upper coloured circle to the corresponding lower circle while avoiding blue obstacle regions. Starting from the vanilla InterGen output, we progressively add penalties that address the initial and final positions, overlap between individuals, intrusion into obstacle regions, and the facing orientation in the final frame, as detailed in the supplementary material.

For each penalty configuration, we generate 24 motion sequences. A sequence is regarded as violating the constraints if either centroid deviates by more than 20 cm, any overlap occurs, an obstacle is penetrated by more than 10 cm, or the final facing directions differ by more than 20 degrees. Table~\ref{tab:motion_constraints_ablation} presents the violation rates: each added penalty reduces its corresponding error, and applying all four yields the lowest violations across all metrics. Qualitative results in Fig.~\ref{fig:ablation} confirm these improvements, progressively changing start/end positions to desired locations (red), removing overlap (green), avoiding obstacles (yellow), and adjusting orientations (blue).

\begin{table}[t]
    \centering
    \resizebox{\columnwidth}{!}{
        \begin{tabular}{lcccc}
            \toprule
            Method & Pos. Err.$\downarrow$ & Overlap$\downarrow$ & Reg. Viol.$\downarrow$ & Orient. Err.$\downarrow$  \\
            \midrule
            InterGen & 1.0 & 0.292 & 0.500 & 1.0 \\
            + $\mathcal{L}_{\text{root}}$ & \textbf{0.0} & 0.333 & 0.917 & 1.0 \\
            + $\mathcal{L}_{\text{overlap}}$ & \textbf{0.0} & \textbf{0.0} & 0.958 & 1.0 \\
            + $\mathcal{L}_{\text{region}}$ & 0.083 & \textbf{0.0} & \textbf{0.043} & 1.0 \\
            + $\mathcal{L}_{\text{orientation}}$ & 0.083 & 0.043 & 0.083 & \textbf{0.208} \\
            \bottomrule
        \end{tabular}
    }
    \vspace{-8pt}
    \caption{Ablation study on motion penalties. Metrics include positional errors (Pos. Err.$\downarrow$), overlap (Overlap$\downarrow$), region violations (Reg. Viol.$\downarrow$), and orientation errors (Orient. Err.$\downarrow$). Positional errors are measured with a threshold of 20 cm, region violations with a threshold of 10 cm, and orientation errors with a threshold of 20 degrees. Lower values indicate better performance.}
    \vspace{-10pt}
    \label{tab:motion_constraints_ablation}
\end{table}

\subsection{Motion Extension}

We evaluate PINO’s ability to generate long-duration motions while maintaining consistency and realism. Using a fixed text prompt, we generate motion sequences and assess them with non-semantic quality metrics. 
As shown in~\Tref{tab:two_person_extension_comparison}, PINO effectively reduces motion artifacts, achieving lower Foot Skating and Max Acc. values than baseline methods, producing smoother and more natural dynamics. 
We further conduct semantic evaluation by aligning generated sequences with ground truth motion. Specifically, we crop generated motions to match the ground truth length and analyze key metrics, including FID, overlap, foot skating, and acceleration. Table~\ref{tab:motion_extension_fid} shows that PINO consistently improves FID while reducing artifacts, ensuring spatial and temporal coherence. The extension, along with the proposed penalties, enables additional variety of interactions. The alternating handshake among three individuals shown in Fig.~\ref{fig:method} is achieved by switching the prompt among two pairs when extending the interactions. Details of the experiments 
are provided in the supplementary material.

\begin{table}[t]
    \centering
    \resizebox{\columnwidth}{!}{
        \begin{tabular}{lccccccc}
            \toprule
            Method &  Overlap↓ & PenVol.↓ & Foot Skate↓ & Max Acc.↓  \\
            \midrule
            InterGen (inpainting)&  0.672 & 15622.16 & 0.070 & 0.044 \\
            PINO-InterGen &  \textbf{0.031} & \textbf{1375.24} & \textbf{0.045} & \textbf{0.043} \\
            \bottomrule
        \end{tabular}
    }
    \vspace{-8pt}
    \caption{Evaluation of non-semantic elements in motion extension.}
    \vspace{-10pt}
    \label{tab:two_person_extension_comparison}
\end{table}

\begin{table}[t]
    \centering
    \resizebox{\columnwidth}{!}{
        \begin{tabular}{lcccc}
            \toprule
            Method & FID↓ & Overlap↓ & Foot Skate↓ & Max Acc.↓  \\
            \midrule
            InterGen  & 20.390 & 0.089 & 0.081 & 0.036 \\
            PINO-InterGen  & \textbf{18.930} & \textbf{0.0} & \textbf{0.054} & \textbf{0.025} \\
            \bottomrule
        \end{tabular}
    }
    \vspace{-7pt}
    \caption{Evaluation of semantic elements in two-person motion extension. Extended motions are cropped to match the ground truth length for evaluation.}
    \vspace{-15.7pt}
    \label{tab:motion_extension_fid}
\end{table}

%% file: sec/6_conclusion.tex
\section{Limitations}
The performance of PINO relies on the quality of the two-person base model, and its focus on pairwise interactions may not fully capture the dynamics of coordinated group behaviors. For example, a person pushing another person into a third person will not be modeled by the underlying two-person model. However, as our framework is able to utilize multiple-person generation method as the base model, future work will focus on enhancing the expressiveness of the base model to better represent higher-order interactions among many individuals and optimizing computational efficiency to support real-time applications.

As the current penalties are based on joint positions, there are some penetrations mainly occurring in regions near the hands, as can be seen in the dance sequence in Fig.~\ref{fig:multi_person_comparison}. This is due to the current interaction dataset missing hand joint information. Adding buffer to the distance penalty to emulate body thickness, or preparing interaction dataset with hand data is required to address the issue. 

\section{Conclusion}
PINO is a training-free framework for generating realistic multi-person interactions without dedicated multi-person training data. PINO extends a pre-trained two-person diffusion model to an arbitrary number of individuals through sequential motion composition and noise optimization, which addresses artifacts like overlap and body penetration for smooth, coherent interactions. PINO also enables motion extension through inpainting techniques, enabling altering roles during interactions. Our approach achieves state-of-the-art performance in large groups and diverse interactions, delivering high-quality results in complex scenarios.

%% file: sec/X_suppl.tex
\clearpage
\maketitlesupplementary
\renewcommand{\thesection}{\Alph{section}}
\setcounter{section}{0}

\section{Formulation of Penalty Functions}
\label{sec:penalty_functions}

\subsection{Root Position Penalty}

To ensure that an individual reaches a specific location at a particular time, we define the root position penalty $\mathcal{L}_{\text{root}}$ as follows:
\begin{equation} \mathcal{L}_{\text{root}} = \sum_{n \in \mathcal{N}} \max\left(0, \left\| \mathbf{p}_{\text{root}}^p(n) - \mathbf{p}_{\text{target}}(n) \right\|^2 - \delta\right), \end{equation} where $\mathbf{p}_{\text{root}}^p(n)$ represents the root position of individual $p$ at frame $n$, and $\mathbf{p}_{\text{target}}(n)$ is the target position at the same frame. The hyperparameter $\delta$ acts as a threshold distance, below which no penalty is applied.

This modification relaxes the loss function by penalizing deviations from the target position only when the distance exceeds the threshold $\delta$. This approach prevents overly rigid constraints on positions that are close enough to the target, enabling more natural and flexible motion generation.

\subsection{Movement Region Penalty}

To restrict movement within a defined area, we introduce the movement region penalty $\mathcal{L}_{\text{region}}$ for a chosen individual $p$:
\begin{equation}
\mathcal{L}_{\text{region}} = \frac{1}{N} \frac{1}{J} \sum_{n} \sum_{j} \phi\left( \mathbf{p}_j^p(n) \right),
\end{equation}
where $N$ is the total number of frames, $J$ is the total number of joints, and $\phi(\mathbf{p})$ penalizes positions outside the desired region. Here, $\mathbf{p}_j^p(n)$ represents the position of joint $j$ of individual $p$ at frame $n$. 

The penalty function $\phi(\mathbf{p})$ can be flexibly defined according to the specific requirements of the task. For example, to restrict movement within a rectangular cuboid region defined by lower bounds $\mathbf{l} = (l_x, l_y, l_z)$ and upper bounds $\mathbf{u} = (u_x, u_y, u_z)$, $\phi(\mathbf{p})$ can be formulated as:
\begin{equation}
\phi\left( \mathbf{p} \right) = \sum_{k \in \{x, y, z\}} \max\left( 0, l_k - p_k \right) + \max\left( 0, p_k - u_k \right),
\end{equation}
where $p_k$ is the $k$-th coordinate of the position $\mathbf{p}$, and $l_k, u_k$ are the lower and upper bounds of the region along the $k$-axis, respectively.

\subsection{Orientation Penalty}

To control facing directions at specific frames while allowing for small deviations, we define the orientation penalty $\mathcal{L}_{\text{orient}}$ with a threshold $\delta$ as follows:
\begin{equation}
\mathcal{L}_{\text{orient}} = \sum_{n \in \mathcal{N}} \max\left(0, 1 - \mathbf{d}^p(n) \cdot \mathbf{d}_{\text{target}}(n) - \delta \right),
\end{equation}
where $\mathbf{d}^p(n)$ is the normalized facing direction of individual $p$ at frame $n$, and $\mathbf{d}_{\text{target}}(n)$ is the desired direction. The hyperparameter $\delta$ acts as a threshold, below which no penalty is applied.

This modification relaxes the loss function by penalizing deviations from the desired direction only when the cosine similarity between $\mathbf{d}^p(n)$ and $\mathbf{d}_{\text{target}}(n)$ falls below $1 - \delta$. By avoiding penalties for minor deviations, this approach enables more natural and flexible motion generation while maintaining alignment with the desired direction.

\subsection{Relative Position Penalty}

To ensure that the root positions of two individuals remain within a desired distance range, 
we define the relative position penalty $\mathcal{L}_{\text{relative}}$ as:
\begin{align}
\mathcal{L}_{\text{relative}} = 
\sum_{n \in \mathcal{N}} \Big[ & 
\max\big(0, d_{\text{min}} - \| \mathbf{p}_{\text{root}}^1(n) - \mathbf{p}_{\text{root}}^2(n) \| \big) \nonumber \\
& + \max\big(0, \| \mathbf{p}_{\text{root}}^1(n) - \mathbf{p}_{\text{root}}^2(n) \| - d_{\text{max}} \big) 
\Big],
\end{align}
where $\mathbf{p}_{\text{root}}^1(n)$ and $\mathbf{p}_{\text{root}}^2(n)$ represent the root positions of 
individuals 1 and 2 at frame $n$, and $d_{\text{min}}$ and $d_{\text{max}}$ are the minimum and maximum 
allowable distances, respectively.

This penalty constrains the distance between the root positions of two individuals to remain within the desired range $[d_{\text{min}}, d_{\text{max}}]$. By adjusting this range, the strictness of the constraint can be flexibly controlled to allow for tighter or more relaxed interactions. The use of $\max$ ensures that penalties are applied only when the distance violates these bounds, enabling natural and adaptable motion generation without imposing overly rigid constraints.

\section{Experimental Details}

\subsection{Optimization Details}

In all experiments, we follow a consistent set of configurations to ensure fair and reproducible results. We base our optimization framework on ProgMoGen~\cite{liu2024programmable}, originally designed for single-person motion optimization. Specifically, we use the Adam optimizer due to its robustness and efficiency in handling motion optimization tasks. The number of optimization steps is set to 100 by default; however, we employ an early stopping criterion that halts optimization once the total loss falls below \(10^{-6}\). This efficiency improvement allows simpler tasks, such as overlap avoidance, to converge more quickly.

For penalty terms, we use fixed hyperparameters throughout all experiments. The overlap penalty threshold, $\delta_{\text{overlap}}$, is set to 30\,cm to ensure sufficient spatial separation. For orientation constraints, we set the threshold $\delta_{\text{orientation}}$ to 0.2. These settings were chosen to balance optimization precision and computational efficiency. The penalty terms introduced in each experiment is shown in Table~\ref{tab:constraints}.

\subsection{Evaluation Metrics}

For the \textit{Overlap} metric, we define an overlap occurrence as any frame in the motion sequence where the root positions of any pair of individuals are closer than 25\,cm. We compute the proportion of generated interactions exhibiting overlaps across all samples, ensuring that even minor instances of unnatural proximity are accounted for. This metric prioritizes generating motions with clear spatial separation between individuals.

\begin{table}[t]
\centering
\resizebox{\columnwidth}{!}{%
\begin{tabular}{lccc}
\toprule
\textbf{Task} & \textbf{Overlap} & \textbf{Root Position} & \textbf{Acceleration} \\ 
\midrule
Two-person (Table 2)         & $\checkmark$ &    &  \\
Multi-person (Table 3, 4, C)       & $\checkmark$    &     &  \\
Two-person Extension (Table 6, 7)          & $\checkmark$    &     & $\checkmark$ \\
Specifying the position (Table B) & $\checkmark$   & $\checkmark$    &  \\
\bottomrule
\end{tabular}%
}
\vspace{-7pt}
\renewcommand{\thetable}{A}
\caption{Penalties applied in each task.}
\vspace{-7pt}
\label{tab:constraints}
\end{table}

\subsection{Tasks}

\subsubsection{Multi-Person Interaction Generation}

The multi-person interaction generation task evaluates the framework's ability to generate plausible interactions between three or more individuals. Initially, two-person interactions are generated based on the text prompt \textit{``the other person approaches one by walking.''} From these interactions, we create eight distinct scenarios. For each scenario, we generate the interaction of a third person 12 times using the same text prompt. During the third person's optimization, we only apply the overlap penalty to guide motion generation, avoiding spatial collisions. This process results in a total of 96 generated interactions, which are quantitatively evaluated using the overlap metric to measure the effectiveness of the penalty as shown in Table~\ref{tab:multi_person_comparison}.

To assess the semantic correctness of the generated interactions, we decompose multi-person interactions into two-person interactions and evaluate them separately. Specifically, we measure FID, overlap, foot skating, and maximum acceleration for each generated pair, where additional individuals are incrementally generated using the first person as a pivot.  Since datasets involving more than three individuals are scarce, we use the same dataset as our two-person interaction evaluation. Specifically, we randomly select 300 samples from the InterGen test set to generate motions and compare them with the ground truth data, applying both quantitative metrics and qualitative analysis. This analysis allows us to quantify how well our method preserves interaction realism while avoiding spatial artifacts. The results are presented in Table~\ref{tab:pairwise_multi_person}, demonstrating that our approach effectively reduces overlap while maintaining natural motion characteristics.

Figure~\ref{fig:multi_person_comparison} illustrates three examples of multi-person interaction generation. In the first example, a two-person dance interaction is generated using the prompt "The two people are dancing together". In the second example, a three-person dance interaction is generated. The initial two-person motion is created using the prompt "Two persons danced at the party". A third person's motion is then generated using the person-to-person approach, optimized with the same prompt and guided by an overlap penalty to avoid spatial collisions.
In the third example, a multi-person conversation scene is generated. First, the motions of two people are created with the prompt “They are talking and using hand gestures.” This two-person interaction is optimized with both an overlap penalty and a root-position penalty to keep each character in its intended area and prevent collisions. One of these two characters is then designated as a hub, and the motions of three additional people are generated sequentially via the person-to-person approach using the same prompt. Each new character is likewise optimized with the overlap penalty and root-position penalty, resulting in a coherent five-person conversation where everyone stands in their assigned locations, gestures naturally, and avoids intersecting with one another.

\subsubsection{Motion Extension}

For the motion extension task in Table~\ref{tab:two_person_extension_comparison}, we first generate two-person interactions lasting 8 seconds based on the text prompt \textit{``Two people are dancing together.''} From each sequence, the final 50 frames are fixed, and an additional 190 frames are generated as a continuation. To ensure smooth transitions and natural motions, we apply two penalties during optimization. The first penalty, the overlap penalty, prevents collisions between individuals during the extended motion. The second penalty, the acceleration penalty, is applied over the first 25 frames of the extended sequence starting from the transition point to suppress abrupt motions. The weights of all penalties and the settings of all hyperparameters are consistent with those used in other experiments. As elements such as abrupt sliding are difficult to observe solely from still images, qualitative results for this task are included in the supplementary video.

For further evaluation, we compare our motion extension results to ground truth sequences by cropping the generated motions to match the ground truth length. This allows for a fair comparison using quantitative metrics, including FID, overlap, foot skating, and maximum acceleration. As shown in Table~\ref{tab:motion_extension_fid}, our method consistently yields improved FID and reduced motion artifacts, demonstrating the capability of our approach to generate smoother and more natural extended motions.

\section{Additional Analysis}

\begin{table*}[t]
    \centering
    \resizebox{\textwidth}{!}{
        \begin{tabular}{l|ccccc|ccc}
            \toprule
            Method & C.Err↓ & Overlap↓ & PenVol.↓ & Foot Skate↓ & Max Acc.↓ & FID↓ & Div.→ & R-Prec.↑ (Top3) \\
            \midrule
            InterGen & 4.029 & 0.119 & 3112.72 & 0.124 & \textbf{0.034}  & 13.278 & 7.793 & 0.674 \\
            InterGen (inpainting) & \textbf{0.0} & 0.130 & 1546.06 & 0.120 & 0.191 & \textbf{12.903} & \textbf{7.899} & \textbf{0.705} \\
            PINO-InterGen & 0.071 & \textbf{0.050} & \textbf{678.32} & \textbf{0.119}  & 0.039 & 14.575 & 7.788 & 0.587 \\
            \bottomrule
        \end{tabular}
    }
    \renewcommand{\thetable}{B}
    \caption{Evaluation of two-person interaction generation by specifying the positions at the initial and final frames.}
    \label{tab:initial_and_final_position_inpainting}
\end{table*}

\begin{figure}[t]
\centering
\begin{subfigure}{0.48\linewidth}
    \centering
    \includegraphics[width=\linewidth]{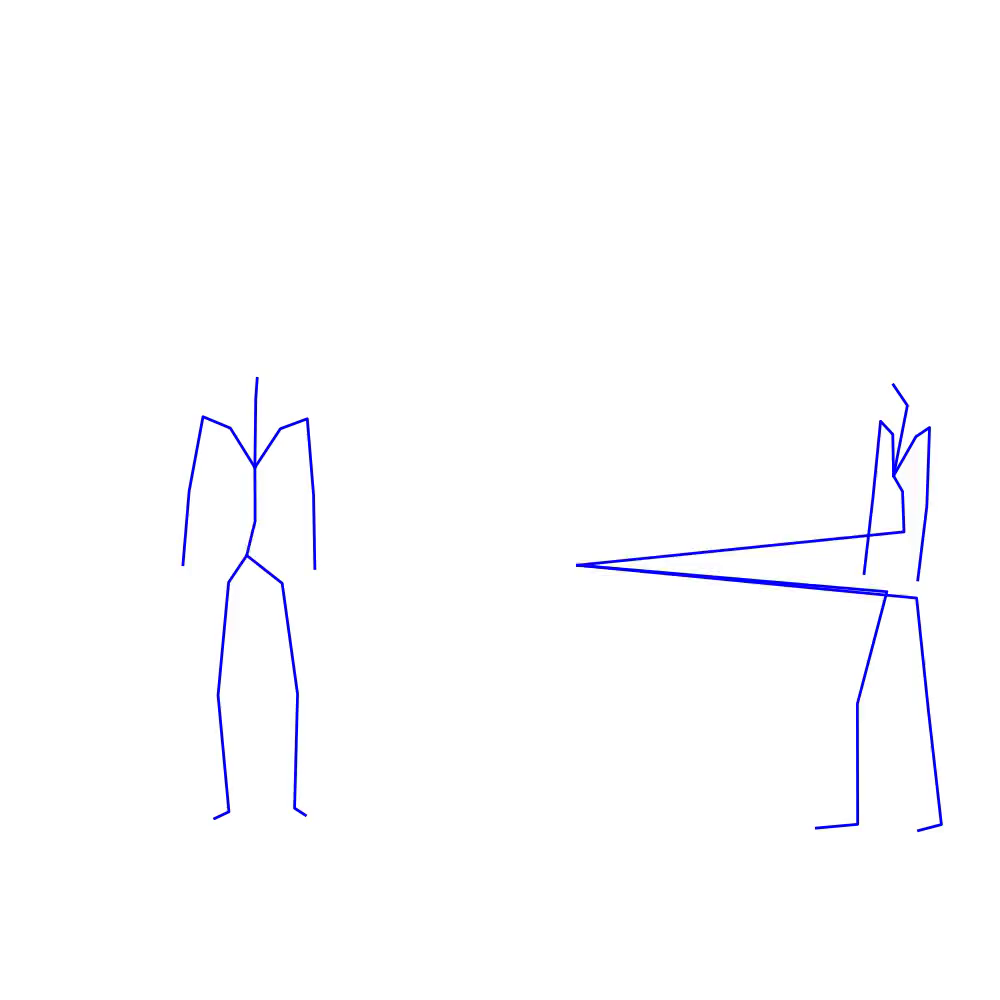}
    
    \label{fig:subfig1}
\end{subfigure}
\hfill
\begin{subfigure}{0.48\linewidth}
    \centering
    \includegraphics[width=\linewidth]{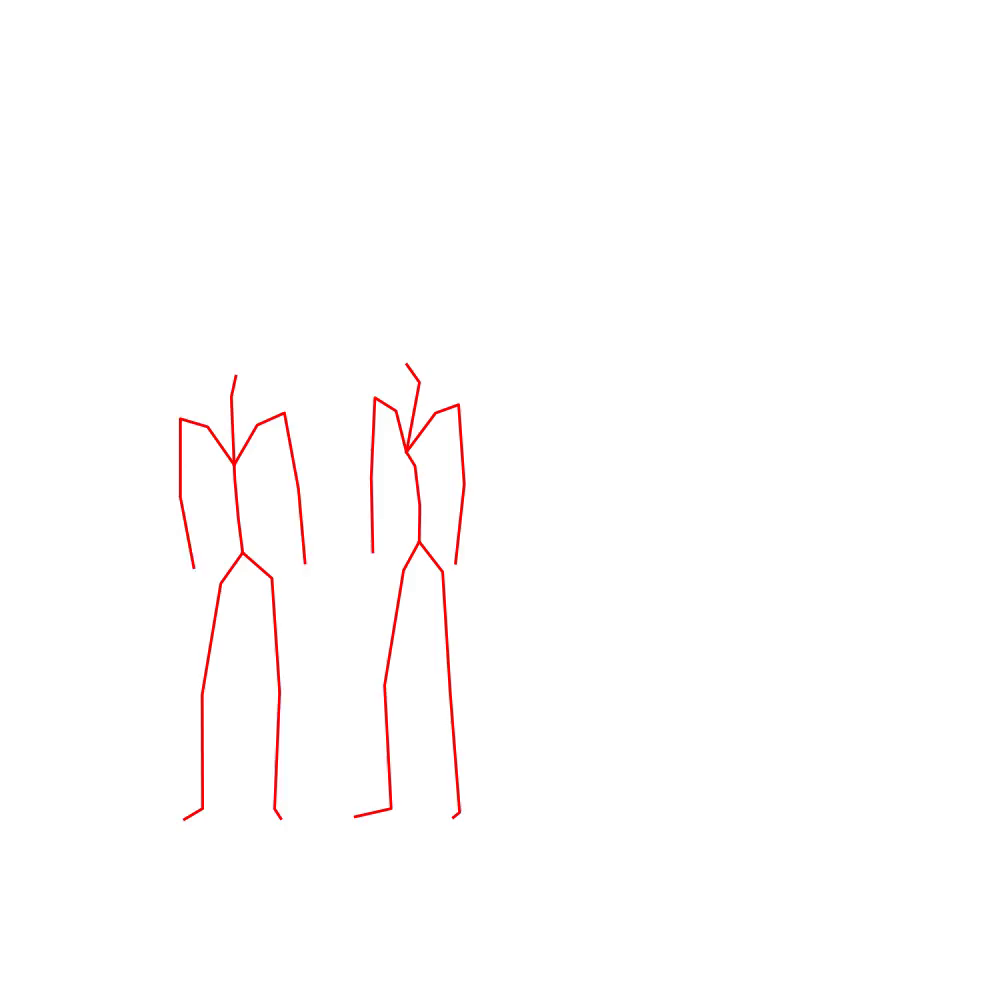} 
    
    \label{fig:initial_frame_comparison}
\end{subfigure}
\renewcommand{\thefigure}{A}
\caption{Initial frames of interaction sequences generated by the inpainting-based method (left) and our optimization-based method (right) in an interaction generation task where root positions are specified at the initial and final frames. The inpainting-based method fails to coordinate the masked, specified root joint with other generated joints, resulting in unnatural skeleton outputs, whereas our method avoids such failures, maintaining consistency with the positional constraints depicted in the initial frames.}
\label{fig:d1}
\end{figure}

\subsection{Comparison with Inpainting-Based Methods}

To further analyze the effectiveness of our approach, we compare it against an inpainting-based method using ground truth (GT) motion data. For the inpainting method, we extract the root positions of the first and final frames from the GT motion data. If the GT sequence exceeds 10 seconds, a random 10-second segment is selected for this process. These root positions are used as constraints, with the corresponding frames masked during generation to guide the motion synthesis. 

As shown in Table~\ref{tab:initial_and_final_position_inpainting}, the inpainting-based method achieves better scores on semantic metrics like FID. However, it exhibits significantly higher values in quantitative metrics such as Max Acc., indicating issues with motion realism. Specifically, as illustrated in Figure~\ref{fig:d1}, the inpainting approach often forces only the root positions of the start and end frames to match the specified constraints, while the relative positions of other joints are misaligned. This results in unnatural motion artifacts. 
In contrast, our method generates plausible motions that adhere to the positional constraints while maintaining realistic joint configurations throughout the sequence. These results highlight the robustness of our approach in preserving the natural structure of motions while fulfilling the specified constraints.

\subsection{Other baseline model}  

\begin{table}
    \centering
    \resizebox{\columnwidth}{!}{
        \begin{tabular}{lccccc}
            \toprule
            Method & Pair & FID↓ & Overlap↓ & Foot Skate↓ & Max Acc.↓  \\
            \midrule
            in2IN & (1,2) & 12.283 & 0.130 & 0.148 & \textbf{0.041}\\
                     & (1,3) & 13.128 & 0.642 & 0.149 & \textbf{0.044}\\
                     & (1,4) & 13.171 & 0.838 & \textbf{0.148} & \textbf{0.046}\\
                     & (1,5) & \textbf{13.161} & 0.920 & \textbf{0.148} & \textbf{0.047}\\
            \midrule
            PINO-in2IN & (1,2) & \textbf{12.183} & \textbf{0.000} & \textbf{0.146} & \textbf{0.041} \\
                         & (1,3) & \textbf{12.811} & \textbf{0.020} & 0.150 &0.045 \\
                         & (1,4) & \textbf{12.967} & \textbf{0.058} & 0.152 &0.049 \\
                         & (1,5) & 13.347 & \textbf{0.103} & 0.153 & 0.051 \\
            \bottomrule
        \end{tabular}
    }
    \vspace{-8pt}
    \renewcommand{\thetable}{C}
    \caption{Evaluation of multi-person interaction generation.}
    \vspace{-10pt}
    \label{tab:pairwise_multi_person_in2IN}
\end{table}

We conducted preliminary comparisons with in2IN~\cite{Ruiz-Ponce_2024_CVPR}, a recent framework for multi-person motion generation. As shown in Table~\ref{tab:pairwise_multi_person_in2IN}, the evaluation follows a similar setting to that of Table~\ref{tab:pairwise_multi_person} in the main paper, where additional individuals are generated incrementally using the first character as a pivot. We observe that in2IN exhibits a comparable trend to InterGen, with increasing overlap as more individuals are introduced. In contrast, our proposed PINO-in2IN significantly reduces such overlap while maintaining competitive or improved performance across FID, Foot Skate, and Max Acc., demonstrating its effectiveness in producing coherent multi-person interactions.

\subsection{User Study}

\begin{figure}
  \centering
  \includegraphics[width=1.0\linewidth]{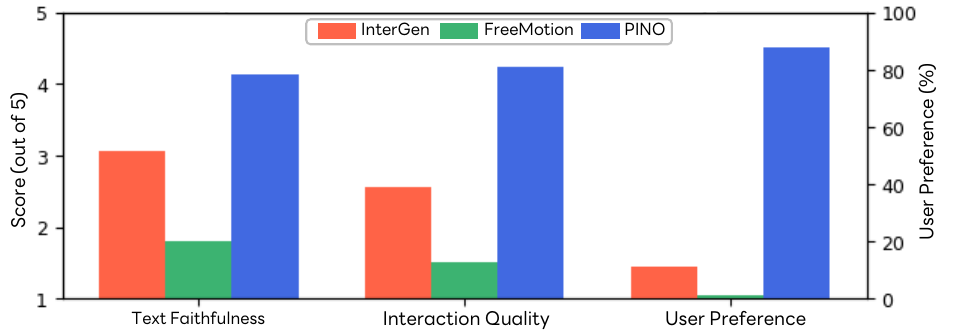}
  \vspace{-20pt}
  \renewcommand{\thefigure}{B}
  \caption{User Study.}
  \vspace{-20pt}
  \label{fig:user_study}
\end{figure}

We conducted a user study with 35 participants to evaluate the quality of motions generated by InterGen~\cite{liang2024intergen}, FreeMotion and ours across five interaction types: ``dance (2/3 person.)'', ``talk'' (Fig.~\ref{fig:multi_person_comparison}), and ``jump,''  ``rock-paper-scissors'' (Fig.~\ref{fig:example}). For each interaction, participants were shown three animations generated from the same text prompt, presented in a randomized order to avoid positional bias. The participants were asked to rate each animation on a 5-point Likert scale for Text Faithfulness and Interaction Quality, which includes physical plausibility factors such as foot contact and absence of overlap. After rating each animation, participants were additionally asked to select the motion they preferred overall for each interaction.

In total, each participant evaluated all five interactions, resulting in 175 evaluation instances. To improve clarity, the animations included ground planes and shadows. As shown in Fig.~\ref{fig:user_study}, our method achieved consistently higher scores in both rating criteria and was most frequently selected in the preference vote.

\subsection{Inference speed}
PINO is optimization based and slower than feed-forward methods. 
On an NVIDIA H100 GPU, inference takes approximately $1$ minute per person for a 10-second motion (300 frames at 30 FPS) with the overlap penalty and early stopping, and up to 10 minutes with additional penalties. While not real-time, it is effective for high-quality offline content of arbitrary-sized group motion, whose training data is scarce. Accelerating optimization is future work.

\subsection{Additional Visualization}
In order to provide better understanding of the effects of our proposal, we include a supplementary video containing the generated results. We highly recommend viewing the video, as the results further emphasize the ability of our method in producing realistic multi-person interactions.